# Optimized Deep Learning Models for AUV Seabed Image Analysis

**Dr. Rajesh Sharma R**
Associate Professor,
Department of Computer Science and Engineering,
Alliance College of Engineering and Design, Alliance University,
Bangalore, India.
rajeshsharma.r@alliance.edu.in,

**Dr. Akey Sungheetha**
Associate Professor,
Department of Computer Science and Engineering,
Alliance College of Engineering and Design, Alliance University,
Bangalore, India.
akey.sungheetha@alliance.edu.in

**Dr. Chinnaiyan R**
Professor,
Department of Computer Science and Engineering,
Alliance College of Engineering and Design, Alliance University,
Bangalore, India.
chinnaiayan.r@alliance.edu.in

*Abstract*: Using autonomous underwater vehicles, or AUVs, has completely changed how we gather data from the ocean floor. AUV innovation has advanced significantly, especially in the analysis of images, due to the increasing need for accurate and efficient seafloor mapping. This blog post provides a detailed summary and comparison of the most current advancements in AUV seafloor image processing. We will go into the realm of undersea technology, covering everything through computer and algorithmic advancements to advances in sensors and cameras. After reading this page through to the end, you will have a solid understanding of the most up-to-date techniques and tools for using AUVs to process seabed photos and how they could further our comprehension of the ocean floor.

## 1 AN OVERVIEW OF AUTONOMOUS UNDERWATER VEHICLE (AUV) SEALED IMAGE PROCESSING

The way we explore and learn about the ocean's depths is completely different with the advent of Autonomous Underwater Vehicles, or AUVs. With their cutting-edge sensors and imaging systems, these unmanned vehicles can take detailed pictures of the underwater environment and seafloor. In several disciplines, including oceanography, geology, marine biology, and underwater archaeology, AUVs are essential. However, processing and analyzing the seabed photos efficiently is hampered by the massive amount of data that AUVs capture. This is where sophisticated image processing methods are useful. Researchers and scientists can gain important insights from these photographs by using advanced algorithms and computational techniques, which will help them better comprehend the oceanographic ecosystems as well as natural structures. Enhancing the quality of the obtained images, eliminating noise and artifacts, and extracting pertinent features for additional analysis are the main goals of AUV seabed image processing. Numerous tasks are involved, including segmentation, object detection, classification, image denoising, and image enhancement. (Source:) Identification and characterization of a wide range of underwater items, including marine life, seafloor sediments, coral reefs, and even man-made structures like shipwrecks, depend on these tasks. Significant progress has been made in AUV seabed image processing in the last several years. To increase the precision and effectiveness of picture analysis, researchers have created cutting-edge methods that make use of computer vision algorithms, deep learning, and machine learning. These developments have made it possible to obtain more precise seafloor mapping, species identification as well as environmental observation. This extensive investigation and comparative research will explore the most recent developments in AUV seabed picture processing. We will examine the many approaches, techniques, and instruments employed in this domain, emphasizing their advantages, drawbacks, and possible uses. When choosing the best image processing techniques for their unique underwater imaging requirements, researchers and practitioners can make well-informed selections by being aware of the state-of-the-art approaches.

Accompany us on this exploratory voyage as we reveal the developments in AUV seabed picture processing and their consequences for underwater study and exploration.

## 2. VALUE OF AUV-SEABED IMAGE PROCESSING IN DIFFERENT SECTORS

AUV seabed image processing is vital to many different businesses; hence its significance cannot be emphasized. The capacity to precisely analyze and interpret seabed images recorded by Autonomous Underwater Vehicles (AUVs) is essential for making informed decisions and optimizing efficiency in a variety of applications, including marine research and exploration, offshore energy production, and underwater infrastructure development. Through non-intrusive means, scientists and researchers can examine and comprehend the marine ecology thanks to AUV seabed image processing in marine research. They can recognize various marine life types, examine their behavior, and evaluate the condition of underwater environments by examining the photos [2]. For the sustainable management of marine resources and conservation initiatives, this knowledge is priceless. AUV seabed image processing is essential to the offshore energy sector's site assessments for renewable energy and oil and gas development projects. AUV-captured photos aid in the planning of the construction of undersea infrastructure, such as pipelines and wind turbines, as well as the identification of possible drilling locations and seabed conditions. By ensuring safe and effective operations and reducing the environmental impact, accurate analysis of these photos is ensured. AUV seabed image processing also has important applications in historical study and underwater archaeology. Through the analysis of the photos, archaeologists can find and record underwater archaeological sites, shipwrecks, and other cultural heritage resources. This information helps to safeguard and preserve these priceless resources in addition to advancing our understanding of the past. Additionally, by aiding habitat mapping and stock evaluation, AUV seabed picture processing assists the fishing sector. Fisheries experts can determine appropriate habitats for sustainable fishing methods, track changes in fish populations, and estimate fish numbers by examining the photos. The long-term sustainability of commercial fisheries is enhanced by this information, which also aids ecosystem-based management strategies. In conclusion, AUV seabed image processing is essential to many different sectors of the economy, such as fisheries management, underwater archaeology, offshore energy generation, and marine research [...]. Accurately analyzing and interpreting these photos offers important insights, supports decision-making, and promotes the sustainable use of the resources found in our oceans.

## 3. RELEASE OF RECENT AUV TECHNOLOGY ADVANCES

Autonomous Underwater Vehicle (AUV) technology has made tremendous strides in the last several years, especially in seabed image processing. The way we explore, and research the ocean's depths has been completely transformed by AUVs, commonly referred to as underwater robots. High-tech sensors and imaging systems on these autonomous vehicles allow them to take detailed pictures of the seafloor. The creation of sophisticated imaging sensors is one of the significant advances in AUV technology [4]. Researchers can now obtain never-before-seen insights into the underwater environment because to these sensors' extraordinary clarity and detail-capable imaging capability. Seabed image accuracy and quality have significantly increased with the use of multi-beam sonar devices and high-definition cameras. AUV navigation and positioning technologies have also undergone substantial advancements. By combining inertial measurement devices with cutting-edge GPS technologies, it is now possible to AUVs to precisely chart their underwater pathways and navigate on their own. This has improved AUV operations' general safety and dependability in addition to increasing data collection efficiency. The field of image processing algorithms has made significant progress as well. Researchers and engineers have created sophisticated algorithms to evaluate and interpret the vast array of complicated and abundant seabed photos that unmanned underwater vehicles (AUVs) are capturing. To identify and categorize objects on the seafloor, these algorithms use strategies such picture segmentation, feature extraction, and pattern recognition. [5] Underwater archeology, environmental monitoring, and marine research have all found this to be quite helpful.

AUV technology has also recently advanced with an emphasis on enhancing the vehicles' durability and energy economy. Because of this, AUVs may now operate for longer periods of time, covering greater ground beneath the surface and gathering more thorough data. AUVs' operational range has been greatly increased by the incorporation of energy harvesting systems and cutting-edge battery technologies, allowing them to carry out longer and more complicated missions. In summary, developments in AUV technology have transformed the analysis of seabed images and significantly increased our comprehension of the undersea environment. Marine research and exploration now have more options because to the development of sophisticated imaging sensors, navigational systems, image processing algorithms, and enhanced endurance. We can anticipate more developments in AUVs as technology progresses, allowing us to explore the ocean's secrets and unearth its hidden riches.

## 4. COMPREHENDING THE DIFFICULTIES IN AUV SEABED IMAGE PROCUREMENT

The use of autonomous underwater vehicles, or AUVs, has completely changed how we investigate and keep an eye on the seabed. High-resolution cameras mounted on these vehicles allow them to take pictures of the underwater environment, which is an invaluable source of information for a variety of uses, including environmental monitoring, marine research, and offshore industry. These photographs of the seabed provide several difficulties in terms of processing and analysis. Poor image quality brought on by things like turbidity, low visibility, and uneven lighting is one of the main problems. Meaningful information extraction is hampered by the noise, blurriness, and distortions that frequently afflict AUV pictures [6].

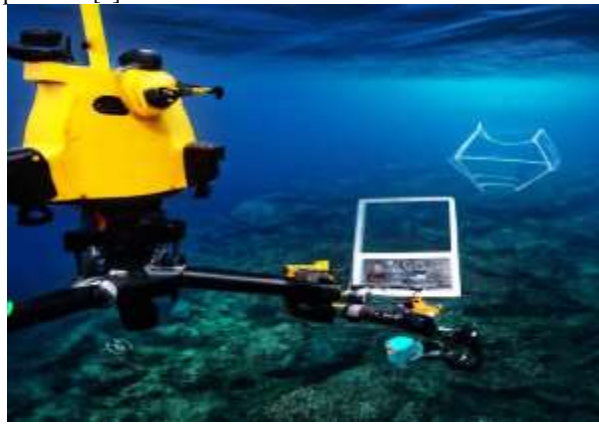

**Figure 1. UAV for Seabed Image Capturing.**

Figure 1 illustrates the difficulty posed by the enormous volume of data that AUVs produce while on mission. These vehicles have the capacity to cover wide regions and take thousands of pictures, which results in an enormous amount of data that needs to be effectively processed and analyzed. To manage the inundation of data and identify pertinent features and patterns, this calls for sophisticated computational approaches and algorithms. The difficulties are further increased by the complexity of the seabed environment. Diverse marine organism species, varied topography, and the presence of detritus, rocks, and corals are characteristics of the seafloor [7]. For the processing algorithms to handle these fluctuations and correctly recognize and classify objects of interest, they must be resilient and flexible.

**Figure 2. Long range UAV for Seabed Image Capturing.**

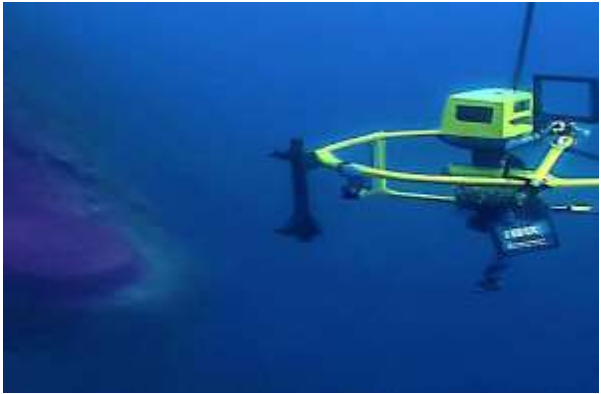

AUV seabed image processing is further complicated by the paucity of ground truth data for training and assessment, as fig. 2 illustrates. In the undersea realm, labeled data for training and testing algorithms is scarcer than for other computer vision applications, such object detection in terrestrial photos. This poses a challenge to the development and validation of efficient image processing methods. In order to overcome these obstacles in AUV seabed picture processing, a thorough grasp of the underlying issues and the development of inventive fixes. To increase image quality, optimize feature extraction, and facilitate precise object detection and classification, scientists, engineers, and other researchers are always investigating novel algorithms, machine learning strategies, and deep learning models.

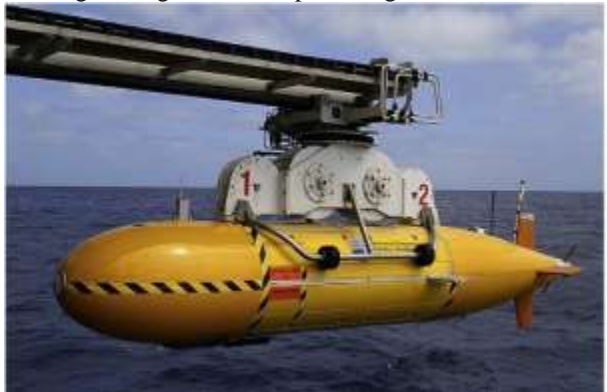

**Figure 3. The UK Natural Environment Council Deepwater rated 6000 meters, Autosub6000 AUV (NERC).**

The UAV that captured the deepest image is shown in Fig. 3 [1]. Advances in AUV seabed image processing have the potential to transform several industries, including underwater archaeology, oil and gas exploration, and marine conservation, if certain obstacles are overcome. We will be able to monitor environmental changes, obtain a deeper understanding of the underwater world, and make well-informed decisions for sustainable resource management thanks to these improvements.

## 5. EXAMINING VARIOUS IMAGE PROCESSING METHODS FOR UNMANNED AUV SENSING

Recent years have witnessed tremendous progress in the field of autonomous underwater vehicles (AUVs), especially in seabed imagery [8]. These developments have produced an abundance of data that may be gathered and examined for a range of purposes, such as offshore resource exploitation, underwater archeology, and environmental monitoring. We will perform a thorough analysis of the various image processing methods that are frequently applied to AUV seabed imaging in this part. Image enhancement is a commonly employed approach in AUV seabed imaging. Using contrast enhancement, sharpness enhancement, and noise reduction, this technique seeks to improve the overall quality of the collected images. In [9] Many strategies have been used to improve the visibility of significant features in the seabed photos, including wavelet-based algorithms, adaptive filtering, and histogram equalization. Image segmentation is another crucial technique that entails breaking an image up into meaningful items or regions. This method is essential to AUV seabed photography to locate and extract particular elements of interest, including underwater flora, coral reefs, or geological formations. To precisely distinguish these features from the background, several segmentation algorithms have been used, such as thresholding, region-based techniques, and edge detection.

AUV seabed imaging also heavily relies on object recognition and detection. Using these methods, particular items or structures in the obtained photos are recognized and categorized. Shipwrecks, pipelines, and marine life are just a few examples of the many seabed items that machine learning techniques, including support vector machines and convolutional neural networks, have shown to be capable of detecting and identifying. [10, 11] Furthermore, the use of 3D reconstruction methods has grown in the field of AUV seabed imaging. Through the combination of many photos captured from various angles, these methods enable the production of precise three-dimensional representations of the seafloor. Underwater navigation, habitat mapping, and volumetric analysis can all be done with this data. To summarize, the examination of several image processing techniques employed in AUV seabed imaging underscores the wide array of instruments and approaches accessible for obtaining significant insights from underwater photography. The progress made in this area has made it possible to map the seabed with more precision and detail, which will help industries, researchers, and scientists make better use of our valuable marine resources.

## 6. COMPARATIVE EVALUATION OF THE EFFICACY AND EFFICIENCY OF IMAGE PROCESSING TECHNIQUES

Ascertaining the efficacy and efficiency of image processing algorithms utilized in AUV seabed investigation is mostly dependent on comparative study. The quality and accuracy of seabed photographs taken by autonomous underwater vehicles (AUVs) are constantly being improved by researchers and engineers working on new algorithms as technology advances. We explore the realm of image processing algorithms in this extensive review, evaluating their effectiveness according to several criteria, including speed, accuracy, computational complexity, and stability under varied seabed circumstances. In [12] The benefits and disadvantages of each algorithm are better understood thanks to this study, which also helps us choose which ones are best for AUV missions. Figure 4 below displays the annual totals of peer-reviewed publications containing fresh AUV-collected marine geoscience data [1]. Assessing the efficacy of algorithms for improving image quality is a crucial component of the comparative study. This covers methods such as picture fusion, edge detection, contrast enhancement, and denoising.

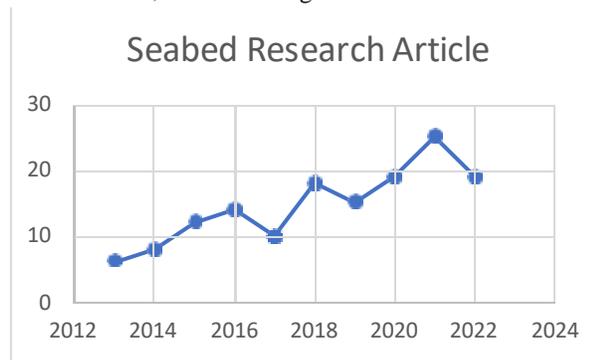

**Figure 4. Graph showing annual totals of peer-reviewed papers featuring new marine geoscience data collected using AUVs during 2013-2022.**

We can identify which algorithms yield the most aesthetically pleasing and educational seafloor photographs by comparing their output. Another important component of AUV image processing is efficiency. The amount of time needed to process and evaluate seabed picture data is strongly impacted by the computational complexity of algorithms. We evaluate the effectiveness of algorithms in terms of processing time and required computer resources through comparison analysis. With the use of this data, engineers and researchers may select algorithms that balance efficiency and accuracy, giving AUVs the ability to process data in real-time or almost real-time.

Moreover, comparative analysis also considers how well algorithms adapt to different bottom circumstances. Unique obstacles arise from different regions and habitats, including variations in seabed topography, water clarity, and the existence of marine life. By observing how algorithms behave in various situations, we can learn more about their resilience and flexibility and choose algorithms that work well under a range of environmental conditions. In general, one of the most important steps toward expanding AUV seabed investigation is the comparative comparison of image processing methods. We can improve the quality of seabed photographs and thereby aid scientific study, resource development, and a better understanding and management of underwater ecosystems by regularly assessing and refining these algorithms.

## 7. CASE STUDIES SHOWCASING EFFECTIVE USES OF AUV SEABED IMAGE PROCESSING

Recent developments in AUV (Autonomous Underwater Vehicle) technology have fundamentally changed how we investigate and comprehend the ocean's depths. Seabed image processing, or the analysis and interpretation of photos taken by underwater cameras, is a crucial part of AUV operations. Examining actual case studies that demonstrate the effective uses of AUV seabed image processing is crucial to fully understanding its potential and significance. [13] These case studies show how this technology has been applied in a variety of fields, including environmental monitoring, marine research, and the exploration of underwater resources.

The application of AUV seabed image processing in marine biodiversity research is demonstrated in one noteworthy case study. An AUV fitted with high-resolution cameras was sent out by researchers to take *pictures of the seafloor and its inhabitants. They improved* our knowledge of undersea ecosystems and their dynamics by identifying and classifying various marine life types by using sophisticated image processing methods [14]. Archaeological surveying is a fascinating use case for AUV seabed picture processing. AUVs with specialized cameras have been used to take close-up pictures of underwater antiquities and historic shipwrecks. By using sophisticated image processing methods, scientists have been able to recreate these ancient artifacts in three dimensions, illuminating our marine history and offering important new perspectives on earlier societies. The analysis of AUV seabed images has been shown to be very helpful in environmental monitoring projects. Scientists can evaluate and analyze changes in marine environments, monitor the health of coral reefs, and spot possible dangers like pollution or invasive species by examining photos taken by AUVs. Strategies for conservation and efficient ecosystem management depend on this knowledge.

Submarine resource mining has greatly benefited from the identification and characterization of highly valuable minerals using AUV seabed image processing. Researchers can minimize environmental damage and cut expenses by identifying regions of interest for more study by assessing the composition and geological formations depicted in the photos. These case studies highlight the enormous potential and many applications of AUV seabed picture processing. We may anticipate more advancements in this area as technology develops, creating new avenues for academic study, commercial use, and preservation of the environment. Scholars and professionals looking to use the potential of AUV seabed image processing in their respective fields will find great value in the thorough assessment and comparative analysis of these developments.

## 8. FUTURE DIRECTIONS AND POSSIBLE IMPROVEMENTS IN AUV SEABED IMAGE PROCESSING

With technology developing at a breakneck speed, the field of AUV seabed image processing has a lot of promises for the future. The capabilities of autonomous underwater vehicles (AUVs) and the precision and effectiveness of seafloor image analysis methods are continually being improved by investigators and researchers who are pushing the envelope. Using artificial intelligence (AI) and machine learning algorithms to AUV seafloor picture processing is one of the future developments that could be possible. AUVs can be trained to identify and categorize various seafloor features more accurately by utilizing AI, which enables more accurate mapping and analysis. The creation of sophisticated imaging sensors and cameras made especially for AUVs is another possible area of progress. These sensors can take detailed pictures and even videos of the seafloor, giving scientists a plethora of important information for deciphering and analyzing. In addition, scientists are investigating the application of cutting-edge computer methods and algorithms to enhance the effectiveness and speed of processing seafloor images. One aspect of this involves the advancement of real-time processing capabilities, which could enable AUVs to assess and decipher photos of the seafloor instantly, facilitating quicker decision-making and action.

Furthermore, the combination of AUVs with other technologies, such satellite imagery and submarine sound systems, creates new avenues for thorough seabed monitoring and research. Through the integration of diverse data sources and processing methodologies, scholars can acquire a more all-encompassing comprehension of the underwater milieu and its dynamics. All things considered, the future of AUV seabed image processing is bright and full of opportunities to improve environmental surveillance, submarine study, and exploration. More effective, precise, and complex methods should be created as science develops, opening the door for fresh discoveries and insights into the enigmatic bottom that constitutes our seas.

## 9. THE EFFECTS OF AUV-SEABED IMAGE ANALYSIS ON CONSERVATION OF NATURE AND SCIENTIFIC RESEARCH

The development of AUV seabed image processing has significantly influenced both scientific inquiry and environmental protection initiatives. Autonomous underwater vehicles (AUVs) with advanced imaging technology have transformed our knowledge of marine environments and the need for conservation since they can take high-resolution pictures of the seabed. The use of AUV seabed image processing techniques has shown to be quite beneficial to scientific investigators. Researchers may identify and analyze a wide range of marine animals, including coral reefs, deep-sea life, and underwater plant life, with never-before-seen detail by examining these photos. More precise species identification, behavior analysis, and habitat mapping are made possible by this degree of visual data, greatly expanding our understanding of ecological diversity in the ocean. Moreover, the analysis of AUV seabed images has shown to be useful in monitoring and evaluating the condition of marine habitats. Researchers and environmental conservationists can gain a better understanding of the effects of human activity on the marine ecosystem by examining changes in seafloor patterns and the presence of pollutants or invasive species. This information helps prevent possible threats to delicate undersea habitats and informs conservation initiatives as well as the construction of marine protected areas. Our comprehension of the effectiveness and possible uses of various AUV seabed image processing methods is improved through a comparative study of these methods. By assessing the effectiveness, precision, and computational efficiency of Investigators can tailor image processing workflows for certain research goals by using a variety of techniques and tools. This guarantees the efficient use of priceless resources, producing outcomes that seem more dependable and perceptive. In conclusion, it is impossible to overestimate the influence that AUV seabed image processing has on the preservation of nature and scientific study. These developments in technology allow scientists to probe farther into the secrets of our seas, revealing the intricacies of marine ecosystems and promoting their preservation. We may work toward a more understanding and ecological approach to preserving our valuable marine resources by consistently enhancing and upgrading these image processing tools.

## 10. THE IMPORTANCE OF ONGOING RESEARCH AND DEVELOPMENT IN AUV SEABED IMAGE PROCESSING, IN SUMMARY

In conclusion, the study of AUV seabed image processing is still developing quickly. It is impossible to exaggerate the importance of ongoing research and development in this field. AUVs have the potential to completely transform several industries, including monitoring the environment, undersea archeology, and oceanic research, as advances in technology occur and our understanding of the undersea world expands. It is clear from this thorough assessment and comparison analysis that scientists and academics are making incredible progress in improving AUVs' capacity for seabed picture processing. When it comes to gathering and interpreting seabed pictures, AUV accuracy, efficiency, and dependability have substantially increased with the introduction of sophisticated algorithms, machine learning and deep learning approaches, and image enhancement methods. It's crucial to remember that there are still issues and constraints that require attention. Accurately processing seafloor photos can be challenging due to various factors such complicated undersea geography, varied seafloor circumstances, and the presence of marine life. Consequently, it will take persistent research and development to get over these challenges and enhance AUVs' capacity for analyzing seabed images. Furthermore, cooperation between scientists, business executives, and government agencies is necessary to guarantee the ethical and sustainable application of AUVs in marine settings. This includes creating uniform procedures, exchanging information and materials, and encouraging moral gathering and analyzing information methods.

To sum up, developments in AUV seabed image processing are revolutionizing our comprehension of the seafloor and creating new avenues for investigation and study. We can anticipate much more advanced AUV systems that will transform numerous sectors and advance our understanding of the Earth's waters thanks to additional study and development. AUV technology is advancing at a rapid pace, and there is a lot of room for growth in this area in the near future.

## REFERENCES

[1] Wynn, R. B., Huvenne, V. A. I., Le Bas, T. P., Murton, B. J., Connelly, D. P., Bett, B. J., Ruhl, H. A., Morris, K. J., Peakall, J., Parsons, D. R., Sumner, E. J., Darby, S. E., Dorrell, R. M., & Hunt, J. E. (2014). Autonomous Underwater Vehicles (AUVs): Their past, present, and future contributions to the advancement of marine geoscience. Marine Geology, 352, 451–468. https://doi.org/10.1016/j.margeo.2014.03.012

[2] Felemban, E., Shaikh, F. K., Qureshi, U. M., Sheikh, A. A., & Qaisar, S. B. (2015). Underwater Sensor Network Applications: A Comprehensive Survey. International Journal of Distributed Sensor Networks, 11(11), 896832. https://doi.org/10.1155/2015/896832

[3] Kocak, D. M., Dalgleish, F. R., Caimi, F. M., & Schechner, Y. Y. (2008). A Focus on Recent Developments and Trends in Underwater Imaging. Marine Technology Society Journal, 42(1), 52–67. https://doi.org/10.4031/002533208786861209

[4] Shihavuddin, A. S. M., Gracias, N., Garcia, R., Gleason, A., & Gintert, B. (2013). Image-Based Coral Reef Classification and Thematic Mapping. Remote Sensing, 5(4), 1809–1841. https://doi.org/10.3390/rs5041809

[5] Donovan, G. T. (2012). Position Error Correction for an Autonomous Underwater Vehicle Inertial Navigation System (INS) Using a Particle Filter. IEEE Journal of Oceanic Engineering, 37(3), 431–445. https://doi.org/10.1109/joe.2012.2190810

[6] Althaus, F., Hill, N., Ferrari, R., Edwards, L., Przeslawski, R., Schönberg, C. H. L., Stuart-Smith, R., Barrett, N., Edgar, G., Colquhoun, J., Tran, M., Jordan, A., Rees, T., & Gowlett-Holmes, K. (2015). A Standardised Vocabulary for Identifying Benthic Biota and Substrata from Underwater Imagery: The CATAMI Classification Scheme. PLOS ONE, 10(10), e0141039. https://doi.org/10.1371/journal.pone.0141039

[7] Hossain, M. S., Bujang, J. S., Zakaria, M. H., & Hashim, M. (2014). The application of remote sensing to seagrass ecosystems: an overview and future research prospects. International Journal of Remote Sensing, 36(1), 61–114. https://doi.org/10.1080/01431161.2014.990649

[8] Raineault, N. A., Trembanis, A. C., & Miller, D. C. (2011). Mapping Benthic Habitats in Delaware Bay and the Coastal Atlantic: Acoustic Techniques Provide Greater Coverage and High Resolution in Complex, Shallow-Water Environments. Estuaries and Coasts, 35(2), 682–699. https://doi.org/10.1007/s12237-011-9457-8

[9] Caimi, F. M., Kocak, D. M., Dalgleish, F., & Watson, J. (2008). Underwater imaging and optics: Recent advances. https://doi.org/10.1109/oceans.2008.5289438

[10] deYoung, B., Visbeck, M., de Araujo Filho, M. C., Baringer, M. O., Black, C., Buch, E., Canonico, G., Coelho, P., Duha, J. T., Edwards, M., Fischer, A., Fritz, J.-S., Ketelhake, S., Muelbert, J.-H., Monteiro, P., Nolan, G., O'Rourke, E., Ott, M., Le Traon, P. Y., & Pouliquen, S. (2019). An Integrated All-Atlantic Ocean Observing System in 2030. Frontiers in Marine Science, 6. https://doi.org/10.3389/fmars.2019.00428

[11] Meadows, G. A. (2013). A review of low cost underwater acoustic remote sensing for large freshwater systems. Journal of Great Lakes Research, 39, 173–182. https://doi.org/10.1016/j.jglr.2013.02.003

[12] Lewis, R., Bose, N., Lewis, S., King, P., Walker, D., Devillers, R., Ridgley, N., Husain, T., Munroe, J., & Vardy, A. (2016). MERLIN - A decade of large AUV experience at Memorial University of Newfoundland. https://doi.org/10.1109/auv.2016.7778675

[13] Niu, B., Li, G., Peng, F., Wu, J., Zhang, L., & Li, Z. (2018). Survey of Fish Behavior Analysis by Computer Vision. Journal of Aquaculture Research & Development, 09(05). https://doi.org/10.4172/2155-9546.1000534

[14] Casalino, G., Allotta, B., Antonelli, G., Caiti, A., Conte, G., Indiveri, G., Melchiorri, C., & Enrico Simetti. (2016). ISME research trends: Marine robotics for emergencies at sea. https://doi.org/10.1109/oceansap.2016.7485616